# PCB-Fire: Automated Classification and Fault Detection in PCB

Tejas Khare[1]
School of Electronics and Communication Engineering
MIT World Peace University
Pune, India
tejas.khare99@gmail.com

Vaibhav Bahel[1]
School of Electronics and Communication Engineering
MIT World Peace University
Pune, India
vaibhavbahel@gmail.com

Anuradha C. Phadke[1]
School of Electronics and Communication Engineering
MIT World Peace University
Pune, India
anuradha.phadke@mitwpu.edu.in

*Abstract*— Printed Circuit Boards ("PCB") are the foundation for the functioning of any electronic device, and therefore are an essential component for various industries such as automobile, communication, computation, etc. However, one of the challenges faced by the PCB manufacturers in the process of manufacturing of the PCBs is the faulty placement of its components including missing components. In the present scenario the infrastructure required to ensure adequate quality of the PCB requires a lot of time and effort. The authors present a novel solution for detecting missing components and classifying them in a resourceful manner. The presented algorithm focuses on pixel theory and object detection, which has been used in combination to optimize the results from the given dataset.

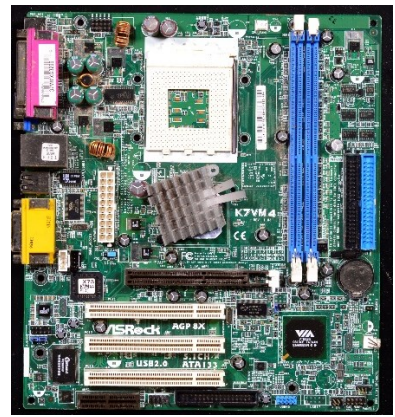
Fig.1 Correct Design PCB

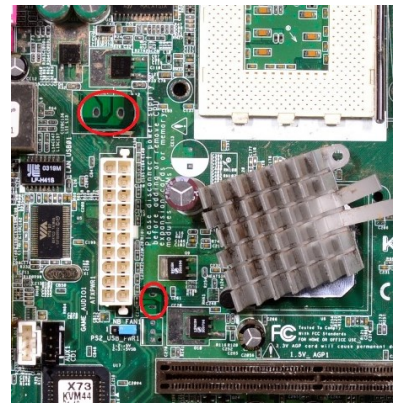
Fig. 2 Component missing PCB (Cropped)

I. INTRODUCTION

*A. Background*

The Printed Circuit Boards (PCB) manufacturing market is a rapidly growing sector and is forecasted to be an $89.7 Billion industry by 2024 [1]. Manufacturing PCB is a very complex and precise process. Often there are cases when the PCBs are manufactured faultily. The most common reasons are 1. Track errors 2. Missing components 3. The orientation of components. The PCBs could be missing some components like a capacitor, a resistor, or an inductor. This fault is caused by either human error or a calibration problem with the assembler. In some cases, the components could be fit in the wrong orientation. These errors significantly affect the performance and life span of the products or the electronic devices in which the PCBs are to be fit and may even lead to PCB failure. This could lead to major problems and even cause the device to collapse.

## B. Literature Survey

An Assembled PCB has the components assembled and soldered by following the design provided to the assembler. Pallavi. M. Taralkar, et al [2] provide a solution for PCB fault detection with the help of image subtraction. The subtracted image is compared to the original PCB design and computes the location of the missing component by classifying it as missing. Jinay Nahar, et al [3], propose an algorithm to inspect faults in assembled PCBs by using image subtraction. The subtracted image is converted to a grayscale image and is passed to an algorithm that determines the faulty regions and classifies them accordingly. The algorithm indicates the position of the fault by drawing a bounding circle. Vikas Krishnaji Salunkhe, et al [4] present a solution to detect faults in PCB by using linear Charged Coupled Device (CCD) PCB images. The features are extracted using pattern recognition techniques, and the faults are identified. S. Sridevi, et al [5] propose a solution to inspect the tracks and layout of a PCB. Their method uses image processing and compares the reference image which has the ground truth design and the test image. If there is a difference between the images, the algorithm classifies it as faulty. Manasa H. R., et al [6] propose a solution to detect a faulty PCB by using the Lab-VIEW tool. The original PCB image and the test image are captured in color format. Furthermore, the captured images are then converted to arrays and compared with each other in a tabular format. The paper uses a binary classification technique to classify the test images as 'failed' or 'passed'. S. H. Indera Putera, et al [7] propose a method to classify bare single layer PCBs as under etch, over etch, short, open circuit, etc. The algorithm uses morphological techniques, image subtraction, image segmentation, and logic gates to find the defects and classify them accordingly. The solution classified 13 defects but with a drawback of generating false-positive images due to noise. The solution can be optimized to classify 14 defects by adding the pin-hole defect as it was ignored due to the morphological image segmentation process. Jithendra P R Nayak, et al [8] present an inspection model to detect faults in a PCB using image processing techniques. The paper uses edge detection, Hough transforms for detecting the faults on single-layer PCBs. The inspection model detects 'cut' in tracks and extra Dry Film Photo Resist (DFPR). The algorithm uses binary conversion to count 0's and 1's to find the region where a fault is present. Yih-Lon Lin, et al [9] present a method for capacitor detection in a PCB using YOLO. The authors trained the YOLOv2 network with 9 different kinds of capacitors.

## II. PROPOSED METHOD

The paper aims to solve the problem of missing components in PCB. A novel solution was presented based on object detection, image subtraction, and pixel manipulation. The authors name this proposed system as *PCB-Fire*. In addition to detecting the missing components, the algorithm also finds its location and its class that is the type of component. The detection and classification of components are achieved by using the YOLO (You Only Look Once) algorithm. The classifier model was trained on a DSLR dataset of 530 PCB images [10]. The detection of missing components is carried out by image subtraction and pixel manipulation.

### A. YOLOv3

It is a neural network used for detection and classification which has free open access [11]. You Only Look Once (YOLO) is a Fully Convolutional Neural Network (FCNN). An input image of t x t size is fed to be classified or detected by the network Fig. 3. The image is split into n x n grids and each box in the grid is visited once by striding. The algorithm produces multiple bounding boxes for a single detection. The best bounding box is selected by Non-Maximum Suppression [12] and others are eliminated. The YOLOv3 does not softmax the class because of the limitation of assuming the classes to be mutually exclusive. Hence, it uses logistic regression and thresholding to predict the classes.

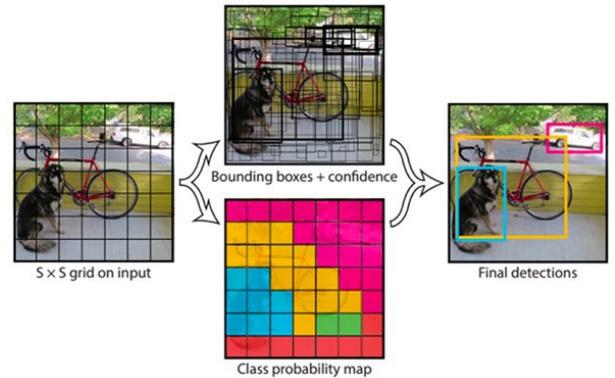

Fig. 3. YOLO algorithm working

The model was trained by specifying the number of filters of the final convolutional layer by the formulae (1). The network consists of a total of 106 CNN layers. There are 75 convolutional layers, 23 shortcut layers, 1 upsample layer, 4 route layers, and finally 3 detection layers that is yolo layers. TABLE I. represents the parameters which were set for the training of the network.

$$filters = (number\ of\ classes + 5) \times 3 \qquad (1)$$

TABLE I. Parameter setting of network

| Parameter | Value |
|---|---|
| Input size x Channel | (608 x 608) x 3 |
| Batch | 24 |
| Subdivisions | 8 |
| Max Epochs | 5000 |
| Optimizer | 'SGD' |
| Learning Rate | 0.001 |
| Momentum | 0.9 |
| Decay | 0.0005 |

### B. Dataset Generation

The object detection model was trained using the open-source DSLR dataset of 530 PCB images [8]. 20 PCB images were taken out for testing. A total of 510 PCB images were used to train the network in which the dataset was split as 90% training samples which are 459 and 10% validating samples which are 51. Table 1. provides the number of individual components used to train the network. The components were labeled using an open-source labeling tool labelImg [13]. The components considered for training the network are Capacitor, Inductor, IC, and Resistor. TABLE II. represents the number of each component considered for training the network.

TABLE II. Categorization of components

| Component | Images/labels passed to the network |
|---|---|
| Capacitor | 2229 |
| Resistor | 1036 |
| Inductor | 282 |
| IC | 2345 |

## C. Algorithm

The design PCB and the test PCB are loaded in binary mode. The images which are loaded in binary are subtracted as shown in Fig. 4 and the subtracted image is thresholded. The pixels which have a color value of 255 are considered and saved temporarily. The original PCB image is sent to the trained object detection model to detect and classify the components using a bounding box. While classifying the components, the data of each bounding box which includes the coordinates of it and the class of the component detected is saved.

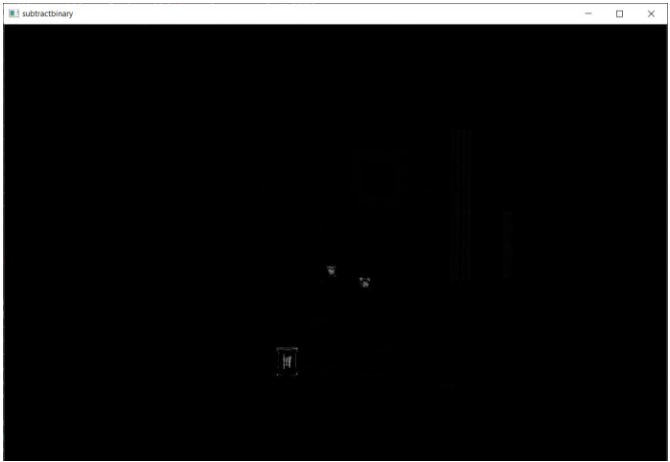

Fig. 4. Binary Subtraction image

Here, this paper introduces a novel algorithm to classify the missing elements of a PCB. At this stage, each bounding box is considered, and the color pixel coordinates obtained earlier are compared with each pixel coordinate residing inside each bounding box. If any pixel coordinate is the same as the color pixel coordinate obtained, a flag is set. After setting the flag, the data of only those boxes are sent to display, whose pixels are matched, i.e. the missing components. The coordinates of the bounding boxes and the class of objects are used to draw the box and display the classification. The authors have named this system as *PCB-Fire*.

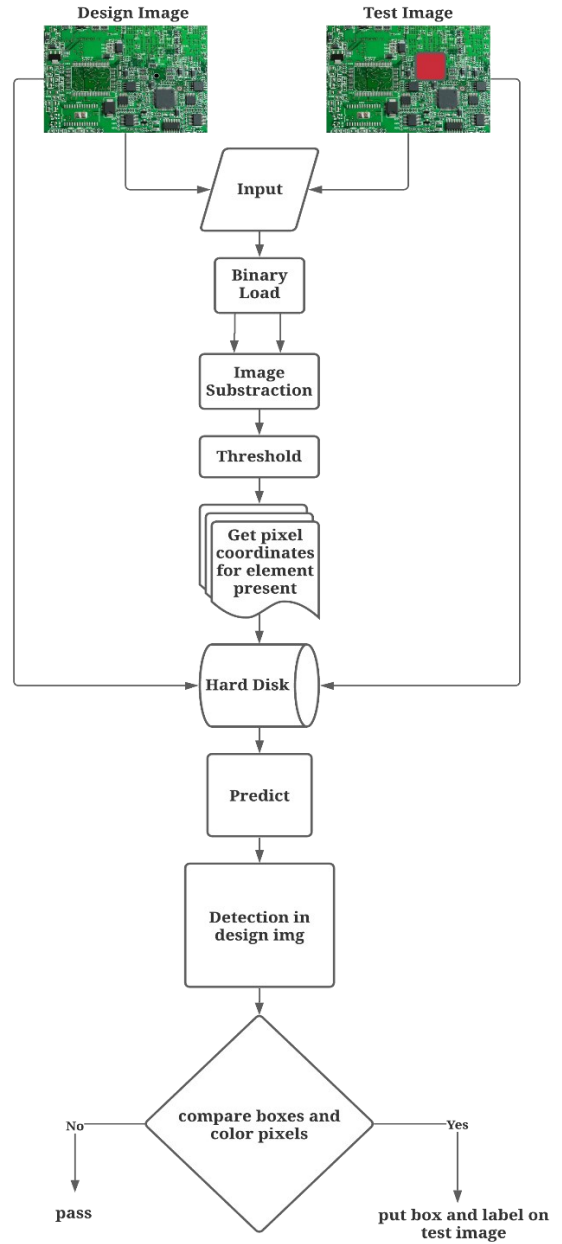

Fig. 5. Flowchart of algorithm

## D. Implementation

1. Take input design image input as des_img and test image as tst_img. Load these images again as a binary image, as variables des_img_b and tst_img_b.
2. Subtract the tst_img_b from the des_img_b for obtaining the binary subtracted image.
   sub_img_b = des_img_b - tst_img_b as shown in Fig. 4

Threshold (sub_img_b, 255 where pixel value greater than average pixel value).
3. Store all pixel coordinates (x, y) in color_pix as sub lists from sub_img_b, where sub_img_b[x][y] == 255. color_pix = [[x1, y1], [x2, y2], [x3, y3], [xn, yn]].
4. Pass des_img, tst_img, color_pix to predict function shown in Fig. 5.
5. The data obtained after the detections and classifications in des_img is saved in idx (idx contains coordinates of each detections) and classIDs (all the labels).
6. Flatten idx, iterate over all boxes using x, y, w, h for each box.
Check each pixel residing inside a bounding box with the color_pix to check any presence of colour pixels. If any pixel coordinates inside the box are matched with the colour pixel coordinate, set flag equal to 1.
for Nth box:
  Flag = 0
  [y_idx, x_idx] = [iterated y coordinate for Nth box, iterated x coordinate for Nth box]
  [x_pix, y_pix] = [iterated sub list from color_pix]
  If for any [y_idx, x_idx] == [x_pix, y_pix]:
    Set Flag = 1

The coordinate axis is inverted with respect to the pixel axis, hence the syntax followed here is (y, x) as shown in Fig. 6

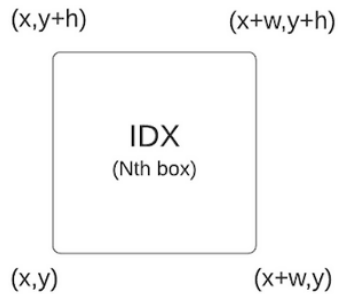

Fig. 6. Coordinates of a bounding box

7. If the flag is equal to 1, draw a bounding box on tst_img.
8. Repeat steps 6 and 7 for getting bounding boxes for all regions where colour pixels are found.
9. Display the tst_img to indicate the missing components and the region.

III. RESULTS

The following results were obtained after testing the system where some of the test cases are shown below. The resultant output images have the detected and classified components as shown in Fig. 7 (a) to Fig. 7 (d) from the yolov3 model. In these results, yellow coloured bounding rectangle is used to represent integrated circuits (IC), green colour is used for capacitor, pink for the inductor, and red for the resistor. Paintbrush tool is used to generate PCB images with missing components which are needed for testing this method. With the paintbrush tool, the bare PCB area from the PCB image is added on a component of the original PCB to create a PCB image with the missing component. Using this technique, the following three sets of Test PCBs were created:

1. Test Set1: PCBs with one missing component
2. Test Set2: PCBs with two missing components
3. Test Set3: PCBs with three missing components

An example of PCB image from Test Set1, Test Set2 and Test Set3 is given in Fig. 8 a), Fig. 8 b), and Fig. 8 c), respectively.

*Note: The output images are cropped and enlarged to showcase the predictions clearly.*

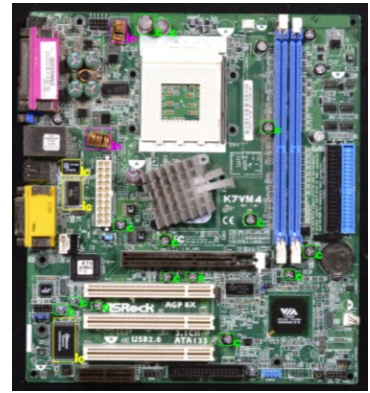

Fig. 7. a) Output 1

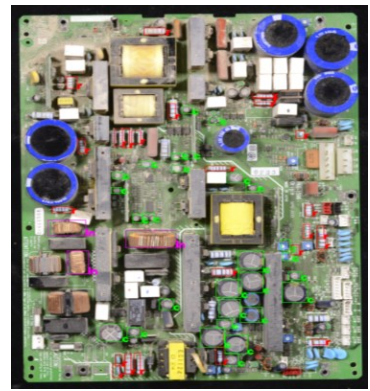

Fig. 7. b) Output 2

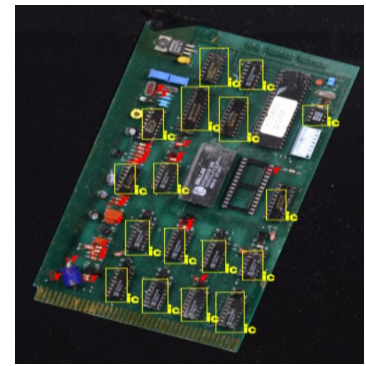

Fig.7. c) Output 3

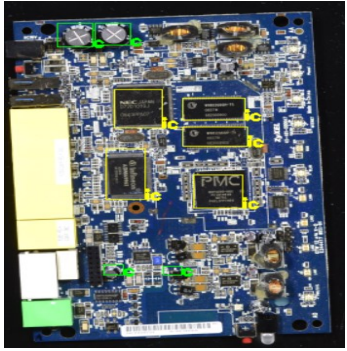

Fig. 7. d) Output 4

Fig. 7. Output of YOLOv3

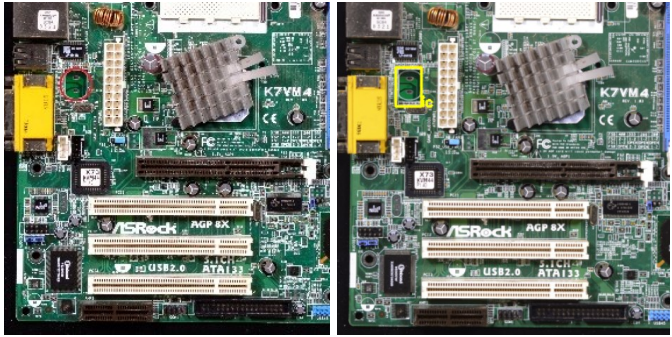

Fig. 8. a) 1 missing component (cropped)

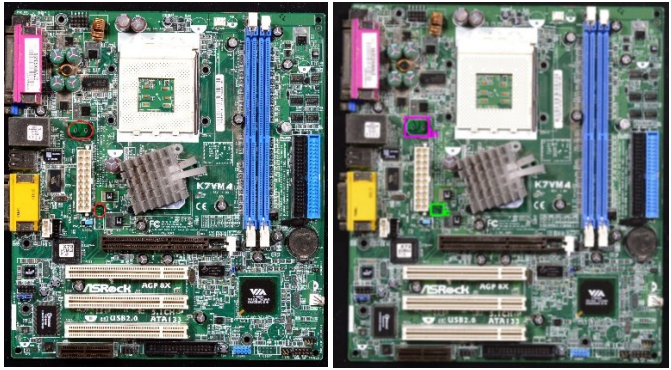

Fig. 8. b) 2 missing components

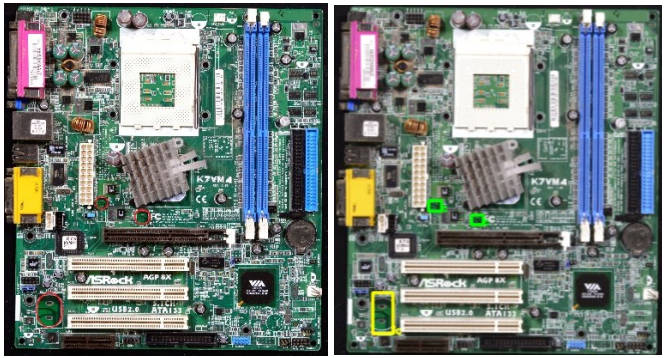

Fig. 8. c) 3 missing components

Fig. 8. Fault detection on test PCB image (missing indicated by a red circle)

## IV. OBSERVATIONS

The algorithm was tested using 20 PCB images in which the set was divided into 1 missing, 2 missing and 3 missing component(s). The validation was also done with PCB images which had different orientations, one of which is shown in Fig. 7 c). The component(s) which are absent, are accurately detected and classified after obtaining the detections and classifications from the object detection model. The algorithm responsible for the detection and classification of the missing component is dependent on the trained object (component) detection model. Moreover, the algorithm correctly displays the missing components and their location as shown in Fig. 8 a) to Fig. 8 c). From Fig. 9, it is observed that the loss at epoch number 4000 was the least, and hence, weights obtained at 4000 epochs were chosen. However, the model was trained up to 5000 epochs, but overfitting was observed after 4000. The loss in Yolo is calculated by using sum of squared errors as shown in (2).

$$SSE = \sum_{i=1}^{n}(x_i - \bar{x})^2 \quad (2)$$

Where n is the number of observations, $x_i$ is the value of the $i^{th}$ observation, and $\bar{x}$ is the mean of all the observations.

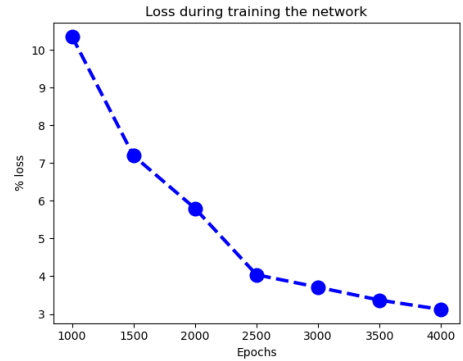

Fig. 9. Validation Loss of network with respect to number of epochs

TABLE III. represents the results obtained from the validation of *PCB-Fire* using a confusion matrix. Where, TP, FP, and FN represent True Positives, False Positives, and False Negatives, respectively.

TABLE III. Results from Test Set

| Components | TP | FP | FN | Total | Accuracy |
|---|---|---|---|---|---|
| Capacitor | 105 | 4 | 18 | 127 | 82.67% |
| Resistor | 39 | 3 | 14 | 56 | 69.64% |
| Inductor | 7 | 3 | 3 | 13 | 53.84% |
| IC | 139 | 2 | 4 | 145 | 95.8% |

The overall accuracy of the algorithm obtained by testing 20 PCB images is 75.48%.

## A. Comparison with existing solutions

The current techniques used for finding missing components only use image-processing. Many researchers [2, 3, 6] present solutions that use image subtraction to carry out missing component detection along with some other image-processing techniques. Whereas, *PCB-Fire* detects the missing components and classifies them using a combination of deep learning, image subtraction and pixel manipulation as its foundation as compared to the method proposed by Y. Lin, et al [9], where only the types of capacitor is being detected.

*PCB-Fire* presents a novel solution as it optimizes the object detection model trained on full board images to work with subtracted images. As the training was done on limited open source images, it lags behind with the detection problem, which the authors solved by pixel manipulation and comparison. The algorithm provides a new insight into the currently used techniques and provides a base to future research projects pertaining to object detection.

## V. CONCLUSION

In this paper, a sublime solution to facilitate the inspection of faulty PCBs was introduced. Furthermore, a novel algorithm using image subtraction and pixel manipulation for detecting and classifying the missing components was proposed. The algorithm was validated by testing multiple images by manually adding bare components on the original PCBs. Considering the long-term losses of machinery and electronics after installing a faulty/unfinished PCB, this solution aims to reduce the loss of capital and time incurred while identifying the fault. The model is giving an accuracy of 75.48%.

## VI. FUTURE SCOPE

This paper intends to present the solution of a frequent day to day problem faced by the PCB manufacturing industry. However, the reach of the method is not limited to just inspecting the missing components and classifying them. The algorithm can be further optimized and extended to inspect problems like misorientation of components, inspecting single layer PCB faults, misplacement of components, i.e. if a wrong component is soldered instead of the actual required one. The inspection of the single layer PCB would involve inspection of tracks which has multiple classes like short, open, and pin-hole defect. Furthermore, the accuracy and speed of component detection can be increased by using YOLOv4 instead of YOLOv3 (which has been used in our method).

## REFERENCES

[1] PR Newswire, July 2019 [online], Available: https://www.prnewswire.com/news-releases/global-printed-circuit-board-pcb-market-expected-to-reach-an-estimated-89-7-billion-by-2024--with-a-cagr-of-4-3-from-2019-to-2024--300879227.html

[2] Pallavi.M.Taralkar, Swati.D.Patil "Image Processing based PCB Component Detection", International Journal of Scientific Research and Review (IJSRR) – Volume 07 Issue 02 - 2018.

[3] Jinay Nahar, Anuradha C. Phadke, "Computer Aided System for Inspection of Assembled PCB", Third International Conference on Intelligent Computing and Control System (ICICCS 2019)

[4] Vikas Krishnaji Salunkhe, Babasaheb Gopal Patil, Suryakant R Dodmise, Neeraj Raavsaheb Patil, "A Review on Study of Fault Detection System for Assembled PCB", International Journal of Engineering Trends and Technology (IJETT) – Volume 34 Number 3- April 2016.

[5] S.Sridevi, G.Muralidharan, C.Nandha Kumar, "Online Inspection of Printed Circuit Board Using Machine Vision", 2014 International Conference on Innovations in Engineering and Technology (ICIET'14).

[6] Manasa H R, Anitha D B, "Fault Detection of Assembled PCB through Image Processing using LABVIEW", International Journal of Engineering Research and Technology (IJERT) – Volume 05 Issue 05 – May 2016.

[7] S.H.Indera Putera, Z.Ibrahim, "Printed Circuit Board Defect Detection Using Mathematical Morphology and MAT LAB Image Processing Tools", 2010 2nd International Conference on Education Technology and Computer (ICETC).

[8] Jithendra P R Nayak, Parameshachari B D, K M Sunjiv Soyjaudah, "Identification of PCB Faults using Image Processing", IEEE 2017 International Conference on Electrical, Electronics, Communication, Computer and Optimization Techniques (ICEECCOT)

[9] Y. Lin, Y. Chiang and H. Hsu, "Capacitor Detection in PCB Using YOLO Algorithm," 2018 International Conference on System Science and Engineering (ICSSE), New Taipei, 2018, pp. 1-4, doi: 10.1109/ICSSE.2018.8520170

[10] PCB DSLR Dataset [online] Available: https://cvl.tuwien.ac.at/research/cvl-databases/pcb-dslr-dataset/

[11] Joseph Redmon, Ali Farhadi "YOLOv3: An Incremental Improvement" [online] Available: https://pjreddie.com/darknet/yolo/

[12] Sambasivarao. K, 'Non-Maximum Suppression', 2019, [online], Available: https://towardsdatascience.com/non-maximum-suppression-nms-93ce178e177c

[13] Tzutalin, LabelImg, [online], Available: https://github.com/tzutalin/labelImg